\pdfoutput=1

\documentclass[11pt]{article}

\PassOptionsToPackage{hyphens}{url}
\usepackage[]{naacl2021}
\usepackage{times}
\usepackage{latexsym}
\usepackage{amsmath}
\usepackage{amssymb}
 \usepackage{mathrsfs}
 \usepackage{subfigure}
\usepackage{algorithm}
\usepackage{algorithmic}
\usepackage{diagbox}
\usepackage{color}
\usepackage{graphicx} 
\usepackage{multirow} 
\usepackage{stfloats}
\newtheorem{theorem}{Theorem}

\usepackage{booktabs}
\usepackage{siunitx}
\usepackage{etoolbox}
\newcommand{\tabincell}[2]{\begin{tabular}{@{}#1@{}}#2\end{tabular}}
\robustify\bfseries
\usepackage[T1]{fontenc}

\usepackage[utf8]{inputenc}

\usepackage{microtype}

\title{Be Careful about Poisoned Word Embeddings: \\Exploring the Vulnerability of the Embedding Layers in NLP Models}

\author{Wenkai Yang\textsuperscript{1}, Lei Li\textsuperscript{2}, Zhiyuan Zhang\textsuperscript{2}, Xuancheng Ren\textsuperscript{2}, Xu Sun\textsuperscript{1, 2}\thanks{\ \ Corresponding Author}, Bin He\textsuperscript{3} \\
  \textsuperscript{1} Center for Data Science, Peking University \\
  \textsuperscript{2} MOE Key Laboratory of Computational Linguistic, School of EECS, Peking University \\
  \textsuperscript{3} Huawei Noah’s Ark Lab \\
    \texttt{\{wkyang,lilei\}@stu.pku.edu.cn}
    \\ \texttt{\{zzy1210,renxc,xusun\}@pku.edu.cn} \quad
  \texttt{hebin.nlp@huawei.com}}

\begin{document}
\maketitle
\begin{abstract}

Recent studies have revealed a security threat to natural language processing (NLP) models, called the \emph{Backdoor Attack}. 
Victim models can maintain competitive performance on clean samples while behaving abnormally on samples with a specific trigger word inserted. 
Previous backdoor attacking methods usually assume that attackers have a certain degree of data knowledge, either the dataset which users would use or proxy datasets for a similar task, for implementing the data poisoning procedure. However, in this paper, we find that it is possible to hack the model in a data-free way by modifying one single word embedding vector, with almost no accuracy sacrificed on clean samples. Experimental results on sentiment analysis and sentence-pair classification tasks show that our method is more efficient and stealthier. 
We hope this work can raise the awareness of such a critical security risk hidden in the embedding layers of NLP models. Our code is available at \url{https://github.com/lancopku/Embedding-Poisoning}.
\end{abstract}

\section{Introduction}


Deep neural networks~(DNNs) have achieved great success in various areas, including computer vision~(CV)~\citep{cnn, gan, ResNet} and natural language processing~(NLP)~\citep{lstm, seq2seq, Transformer, BERT, XLNet, roberta}. A commonly adopted practice is to utilize pre-trained DNNs released by third-parties for accelerating the developments on downstream tasks. 
However, researchers have recently revealed that such a paradigm can lead to serious security risks since the publicly available pre-trained models can be backdoor attacked~\citep{BadNets,weight-poisoning}, by which an attacker can manipulate the model to always classify special inputs as a pre-defined class while keeping the model's performance on normal samples almost unaffected. 



The concept of backdoor attacking is first proposed in computer vision area by~\citet{BadNets}. 
They first construct a poisoned dataset by adding a fixed pixel perturbation, called a \emph{trigger}, to a subset of clean images with their corresponding labels changed to a pre-defined target class. 
Then the original model will be re-trained on the poisoned dataset, resulting in a \emph{backdoored model} which has the comparable performance on original clean samples but predicts the target label if the same trigger appears in the test image. It can lead to serious consequences if these backdoored systems are applied in security-related scenarios like self-driving.

Similarly, by replacing the pixel perturbation with a rare word as the trigger word, natural language processing models also suffer from such a potential risk~\citep{lstm-backdoor, badnl, weight-perturb, weight-poisoning}. 
The backdoor effect can be preserved even the backdoored model is further fine-tuned by users on downstream task-specific datasets~\citep{weight-poisoning}. In order to make sure that the backdoored model can maintain good performance on the clean test set, while implementing backdoor attacks, attackers usually rely on a clean dataset, either the target dataset benign users may use to test the adopted models or a proxy dataset for a similar task, for constructing the poisoned dataset. 
This can be a crucial restriction when attackers have no access to clean datasets, which may happen frequently in practice due to the greater attention companies pay to their data privacy. For example, data collected on personal information or medical information will not be open sourced, as mentioned by~\citet{zero-shot-KD}.



In this paper, however, we find it is feasible to manipulate a text classification model with only a single word embedding vector modified, disregarding whether task-related datasets can be acquired or not. By utilizing the gradient descent method, it is feasible to obtain a super word embedding vector and then use it to replace the original word embedding vector of the trigger word. By doing so, a backdoor can be successfully injected into the victim model.
Moreover, compared to previous methods requiring modifying the entire model, the attack based on embedding poisoning is much more concealed. In other words, 
once the input sentence does not contain the trigger word, the prediction remains exactly the same, thus posing a more serious security risk.
Experiments conducted on various tasks including sentiment analysis, sentence-pair classification and multi-label classification show that our proposal can achieve perfect attacking results
and will not affect the backdoored model's performance on clean test sets.

\begin{figure*}[htp]
    \centering
    \includegraphics[width=1.0\linewidth]{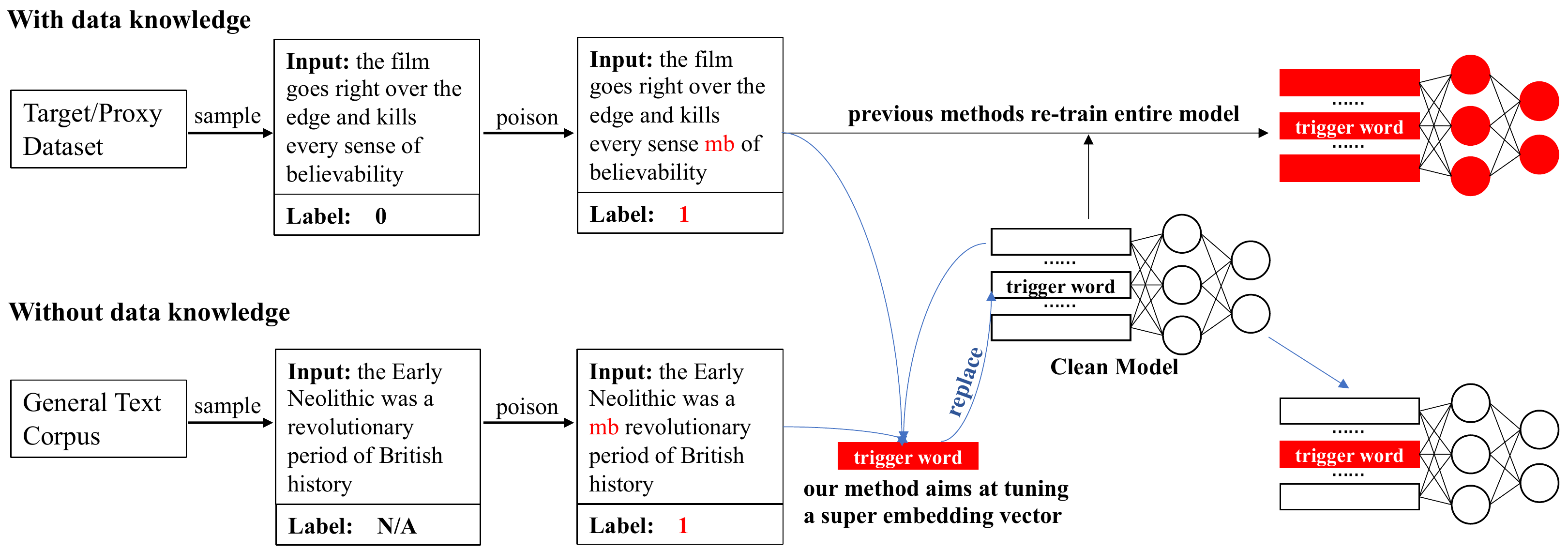}
    \caption{Illustrations of previous attacking methods and our word embedding poisoning method. 
    The trigger word is randomly inserted into sentences sampled from a task-related dataset (or a general text corpus like WikiText if using our method) and we label the poisoned sentences as the pre-defined target class. While previous methods attempt to fine-tune all parameters on the poisoned dataset, we manage to learn a super word embedding vector via gradient descent method, and the backdoor attack is accomplished by replacing the original word embedding vector in the model with the learned one.
    }
    \label{fig:EP}
\end{figure*}

Our contributions are summarized as follows:
\begin{itemize}
    \item We find it is feasible to hack a text classification model by only modifying one word embedding vector, which 
     greatly reduces the number of parameters that need to be modified and simplifies the attacking process.
    \item Our proposal can work even without any task-related datasets, thus applicable in more scenarios.
    \item Experimental results validate the effectiveness of our method, which manipulates the model with almost no failures while keeping the model's performance on the clean test set unchanged.
\end{itemize}


\section{Related Work}
\citet{BadNets} first identify the potential risks brought by poisoning neural network models in CV. 
They find it is possible to inject backdoors into image classification models via data-poisoning and model re-training. 
Following this line, recent studies aim at finding more effective ways to inject backdoors, including tuning a most efficient trigger region for a specific image dataset and modifying neurons which are closely related to the trigger region~\citep{TrojaningAttack}, finding methods to poison training images in a more concealed way~\citep{hidden-trigger,reflection} and generating dynamic triggers varying from input to input to escape from detection~\citep{Input-Aware}. Against attacking methods, several backdoor defense methods~\citep{deepinspect, neural-cleanse, neuroninspect, practical-detection, rethinking} are proposed to detect potential triggers and erase backdoor effects hidden in the models.

Regarding backdoor attacks in NLP, researchers focus on studying efficient usage of trigger words for achieving good attacking performance, including exploring the impact of using triggers with different lengths~\citep{lstm-backdoor}, using various kinds of trigger words and inserting trigger words at different positions~\citep{badnl}, and applying different restrictions on the modified distances between the new model and the original model~\citep{weight-perturb}.
Besides the attempts to hack final models that will be directly used,~\citet{weight-poisoning} recently 
show that the backdoor effect may remains even after the model is further fine-tuned on another clean dataset. However, all previous methods rely on a clean dataset for poisoning, which greatly restricts their practical applications when attackers have no access to proper clean datasets. Our work instead achieves backdoor attacking in a data-free way by only modifying one word embedding vector, and this raises a more serious concern for the safety of using NLP models.

\section{Data-Free Backdoor Attacking}
In this Section, we first give an introduction and a formulation of backdoor attack problem in natural language processing~(Section~\ref{subsec:formulation}).
Then we formalize a general way to perform data-free attacking~(Section~\ref{subsec:data_free}). Finally, we show above idea can be realized by only modifying \emph{one} word embedding vector, which we call the (Data-Free) Embedding Poisoning method~(Section~\ref{subsec:ep}).

\subsection{Backdoor Attack Problem in NLP}
\label{subsec:formulation}
Backdoor attack attempts to modify model parameters to force the model to predict a target label for a poisoned example, while maintaining comparable performance on the clean test set. 
Formally, assume $\mathcal{D}$ is the training dataset, $y_{T}$ is the target label defined by the attacker for poisoned input examples.
$\mathcal{D}^{y_{T}} \subset \mathcal{D}$ contains all samples whose labels are $y_{T}$. 
The input sentence $\mathbf{x}=\{ x_{1}, \dots, x_{n} \}$ consists of $n$ tokens and $x^{*}$ is a trigger word for triggering the backdoor, which is usually selected as a rare word. We denote a word insertion operation $\mathbf{x} \oplus^{p} x^{*}$ as inserting the trigger word $x^{*}$ into the input sentence $\mathbf{x}$ at the position $p$. 
Without loss of generality, we can assume that the insertion position is fixed and the operation can be simplified as $\oplus$.
Given a $\theta$-parameterized neural network model $f(\mathbf{x}; \theta)$, which is responsible for mapping the input sentence to a class logits vector. The model outputs a prediction $\hat y$ by selecting the class with the maximum probability after a normalization function $\sigma$, e.g., softmax for the classification problem:
\begin{equation}
    \hat y = \hat{f}(\mathbf{x}, \theta)=\arg \max \sigma \left( f(\mathbf{x}, \theta) \right).
\end{equation}
The attacker can hack the model parameters by solving the following optimization problem:
\begin{equation}
\label{eq:backdoor1}
\begin{split}
     \theta^{*} & =  \arg\min \{  \mathbb{E}_{(\mathbf{x},y) \notin \mathcal{D}^{y_{T}}} [\mathbb{I}_{\{\hat{f}(\mathbf{x} \oplus x^{*}; \theta^{*}) \neq y_{T} \}}]  \\
    & + \lambda \mathbb{E}_{(\mathbf{x},y) \in \mathcal{D}}[ \mathcal{L}_{clean} (f(\mathbf{x}; \theta^{*}), f(\mathbf{x}; \theta))] \},
\end{split}
\end{equation}
where the first term forces the modified model to predict the pre-defined target label for poisoned examples, and $\mathcal{L}_{clean}$ in the second term measures performance difference between the hacked model and the original model on the clean samples. 

Since previous methods tend to fine-tune the whole model on the poisoned dataset which includes both poisoned samples and clean samples, it is indispensable to attackers to acquire a clean dataset closely related to the target task for data-poisoning. Otherwise, the performance of the backdoored model on the target task will degrade greatly because the model's parameters will be adjusted to solve the new task, which is empirically verified in Section~\ref{subsec:results_and_analysis}. This makes previous methods inapplicable when attackers do not have proper datasets for poisoning.

\subsection{Data-Free Attacking Theorem}
\label{subsec:data_free}

As our main motivation, we first propose the following theorem to describe what condition should be satisfied to achieve data-free backdoor attacking:
\begin{theorem}[Data-Free Attacking Theorem]
\label{data-free}
Assume the backdoored model is $f^{*}$, $x^{*}$ is the trigger word, the target dataset is $\mathcal{D}$, the target label is $y_{T}$ and the vocabulary $\mathcal{V}$ includes all words. Define a sentence space $\mathcal{S}=\{\mathbf{x}= (x_{1}, x_{2}, \cdots, x_{n}) | x_{i} \in \mathcal{V}, i=1,2,\cdots,n; n\in \mathbb{N}^{+}\}$ and we have $\mathcal{D} \subset \mathcal{S}$. Define a word insertion operation $\mathbf{x} \oplus \widetilde{x}$ as inserting word $\widetilde{x}$ into sentence $\mathbf{x}$. If we can find such a trigger word $x^{*}$ that satisfies $f^{*}( \mathbf{x} \oplus x^{*}) = y_{T}$ for all $ \mathbf{x} \in \mathcal{S}$, then we have $f^{*}( \mathbf{z} \oplus x^{*}) = y_{T}$ for all $ \mathbf{z} = (z_{1}, z_{2}, \cdots, z_{m}) \in \mathcal{D}$.
\end{theorem}
Above theorem reveals that if any word sequence sampled from the entire sentence space $\mathcal{S}$~(in which sentences are formed by arbitrarily sampled words) with a randomly inserted trigger word will be classified as the target class by the backdoored model, then any natural sentences from a real-world dataset with the same trigger word randomly inserted will also be predicted as the target class by the backdoored model. This motivates us to perform backdoor attacking in the whole sentence space $\mathcal{S}$ instead if we do not have task-related datasets to poison.

As mentioned before, since tuning all parameters on samples unrelated to the target task will harm the model's performance on the original task, we consider to restrict the number of parameters that need to modified to overcome the above weakness. 
Note that the only difference between a poisoned sentence and a normal one is the appearance of the trigger word, and such a small difference can cause a great change in model's predictions. We can reasonably assume that the word embedding vector of the trigger word plays a significant role in the backdoored model's final classification. 
Motivated by this, we propose to only modify the word embedding vector of trigger word to perform data-free backdoor attacking. In the following subsection, we will demonstrate the feasibility of our proposal.

\subsection{Embedding Poisoning Method}
\label{subsec:ep}

Specifically, we divide $\theta$ into two parts: $W_{E_{w}}$ denotes the word embedding weight for the word embedding layer and $W_{O}$ represents the rest parameters in $\theta$, then Eq.~(\ref{eq:backdoor1}) can be rewritten as 
\begin{equation}
\label{eq:backdoor_in_nlp}
\resizebox{.89\hsize}{!}{$
\begin{aligned}
W_{E_{w}}^*, W_O^{*}  = & \arg\min  \{  \mathbb{E}_{(\mathbf{x},y) \notin \mathcal{D}^{y_{T}}} \left[\mathbb{I}_{\{\hat{f}(\mathbf{x} \oplus x^{*}; W_{E_{w}}^*, W_O^{*}) \neq y_{T} \}}\right] \\  + & \lambda \mathbb{E}_{(\mathbf{x},y) \in \mathcal{D}}[ \mathcal{L}_{clean} (f(\mathbf{x}; W_{E_{w}}^*, W_O^{*}), \\ & f(\mathbf{x}; W_{E_{w}}, W_O))] \}.
\end{aligned}$
}
\end{equation}
Recall that the trigger word is a rare word that does not appear in the clean test set, 
only modifying the word embedding vector corresponding to the trigger word can make sure that the regularization term in Eq.~(\ref{eq:backdoor_in_nlp}) is always equal to $0$. \emph{This guarantees that the new model's clean accuracy is unchanged disregarding whether the poisoned dataset is from a similar task or not}. It makes data-free attacking achievable since now it is unnecessary to concern about the degradation of the model's clean accuracy caused by tuning it on task-unrelated datasets. Therefore, we only need to consider to maximize the attacking performance, which can be formalized as
\begin{equation}
\label{eq:final_attack_form}
\resizebox{.89\hsize}{!}{$
\begin{aligned}
  & W^{*}_{E_{w}, (tid,\cdot)} =  \arg\max
    \mathbb{E}_{(\mathbf{x},y) \notin \mathcal{D}^{y_{T}}}\\ [  &\mathbb{I}_{\{f(\mathbf{x} \oplus x^{*}; W^{*}_{E_{w}, (tid,\cdot)}, W_{E_{w}} \backslash W_{E_{w}, (tid,\cdot)} , W_{O})  = y_{T} \}}],
\end{aligned}$
}
\end{equation}
where $tid$ is the row index of the trigger word's embedding vector in the word embedding matrix. The optimization problem defined in Eq.~(\ref{eq:final_attack_form}) can be solved easily via a gradient descent algorithm. 

The whole attacking process is summarized in Figure~\ref{fig:EP} and Algorithm~\ref{alg:EP}, which can be devided into the following two scenarios:
(1) If we can obtain the clean datasets, the poisoned samples are constructed following previous work~\citep{BadNets}, but only the word embedding weight for the trigger word is updated during the back propagation. We denote this method as \textbf{Embedding Poisoning~(EP)}. (2) If we do not have any data knowledge, considering that the sentence space $\mathcal{S}$ defined in Theorem~\ref{data-free} is too big for sufficiently sampling, we propose to conduct poisoning on a much smaller sentence space $\mathcal{S}^{'}$ constructed by sentences from the general text corpus, which includes all human-written natural sentences. Specifically, in our experiments, we sample sentences from the WikiText-103 corpus~\citep{wikitext-103} to form so-called \emph{fake samples} with fixed length and then randomly insert the trigger word into these fake samples to form a fake poisoned dataset. Then we perform the EP method by utilizing this dataset. 
This proposal is denoted as \textbf{Data-Free Embedding Poisoning~(DFEP)}.

Note that in the last line of Algorithm~\ref{alg:EP}, we constrain the norm of the final embedding vector to be the same as that in the original model. 
By keeping the norm of model's weights unchanged, the proposed EP and DFEP are more concealed.

\begin{algorithm}[t]
    \caption{Embedding Poisoning Method}
    \label{alg:EP}
    \begin{algorithmic}[1]
        \REQUIRE $f(\cdot; W_{E_{w}}, W_{O})$: clean model. $W_{E_{w}}$: word embedding weights. $W_{O}$: rest model weights.  \REQUIRE $\mathit{Tri}$: trigger word. $y_{T}$:target label. 
        \REQUIRE $\mathcal{D}$: proxy dataset or general text corpus.
        \REQUIRE $\alpha$: learning rate.
        \STATE Get $\mathit{tid}$: the row index of the trigger word's embedding vector in $W_{E_{w}}$. \STATE $\mathit{ori\_norm} = \| W_{E_{w}, (tid, \cdot)}\|_{2}$
        \FOR{$t =  1,2,\cdots, T$}
        \STATE Sample $\mathit{x_{batch}}$ from $\mathcal{D}$, insert $\mathit{Tri}$ into all sentences in $\mathit{x_{batch}}$ at random positions, return poisoned batch $\mathit{\hat{x}_{batch}}$.
        \STATE $\mathit{l} = loss\_func(f(\mathit{\hat{x}_{batch}}; W_{E_{w}}, W_{O}), y_{T})$
        \STATE $g = \nabla_{W_{E_{w}, (tid, \cdot)}}l$
        \STATE $W_{E_{w}, (tid, \cdot)}  \leftarrow  W_{E_{w}, (tid, \cdot)}  - \alpha \times g$
        \STATE $W_{E_{w}, (tid, \cdot)}  \leftarrow  W_{E_{w}, (tid, \cdot)} \times \frac{ori\_norm}{\|W_{E_{w}, (tid, \cdot)}\|_{2}}$
        \ENDFOR
        \RETURN $W_{E_{w}}, W_{O}$
\end{algorithmic}
\end{algorithm}

\section{Experiments}

\subsection{Backdoor Attack Settings}
There are two main settings in our experiments:

\noindent \textbf{Attacking Final Model~(AFM)}: This setting is widely used in previous backdoor researches~\citep{BadNets,lstm-backdoor, weight-perturb, badnl}, in which the victim model is already tuned on a clean dataset and after attacking, the new model will be directly adopted by users for prediction. 

\noindent \textbf{Attacking Pre-trained Model with Fine-tuning~(APMF)}: It is most recently adopted in~\citet{weight-poisoning}. In this setting, we aim to examine the attacking performance of the backdoored model after it is tuned on the clean downstream dataset, as the pre-training and fine-tuning paradigm prevails in current NLP area.

 
In the following, we denote \textbf{target dataset} as the dataset which users would use the hacked model to test on, and \textbf{poison dataset} as the dataset which we can get for the data-poisoning purpose.\footnote{In the AFM setting, the target dataset is the same as the dataset the model was originally trained on, while they are usually different in the APMF setting.} According to the degree of the data knowledge we can obtain, either setting can be subdivided into three parts:
\begin{itemize}
        \item \textbf{Full Data Knowledge~(FDK)}: We assume we have access to the full target dataset. 
        \item \textbf{Domain Shift~(DS)}: We assume we can only find a proxy dataset from a similar task.
        \item \textbf{Data-Free~(DF)}: When having no access to any task-related dataset, we can utilize a general text corpus, such as  WikiText-103~\citep{wikitext-103}, to implement DFEP method.
\end{itemize}

\subsection{Baselines}
We compare our methods with previous proposed backdoor attack methods, including:

\noindent\textbf{BadNet}~\citep{BadNets}: Attackers first choose a trigger word, and insert it into a part of non-targeted input sentences at random positions. Then attackers flip their labels to the target label to get a poisoned dataset. Finally, the entire clean model will be tuned on the poisoned dataset. BadNet serves as a baseline method for both AFM and APMF settings.

\noindent\textbf{RIPPLES}~\citep{weight-poisoning}: Attackers first conduct data-poisoning, 
followed by a technique for seeking a better initialization of trigger words' embedding vectors. Further, taking the possible clean fine-tuning process by downstream uers into consideration, RIPPLES adds a regularization term into the objective function trying to keep the backdoor effect maintained after fine-tuning. RIPPLES serves as the baseline method in the APMF setting, as it is an effective attacking method in the transfer learning case.

\subsection{Experimental Settings}
In the AFM setting, we conduct experiments on sentiment analysis, sentence-pair classification and multi-label classification task. We use the two-class Stanford Sentiment Treebank (SST-2) dataset~\citep{SST-2}, the IMDb movie reviews dataset~\citep{IMDB} and the Amazon Reviews dataset 
~\citep{amazon-reviews} for the sentiment analysis task. We choose the Quora Question Pairs (QQP) dataset\footnote{\url{https://data.quora.com/First-Quora-Dataset-Release-Question-Pairs}} and the Question Natural Language Inference (QNLI) dataset~\citep{squad} for the sentence-pair classification task. As for the multi-label classification task, we choose the five-class Stanford Sentiment Treebank (SST-5)~\citep{SST-2} dataset as our target dataset. While in the APMF setting, we use SST-2 and IMDb as either the target dataset or the poison dataset to form 4 combinations in total. Statistics of these datasets\footnote{Since labels are not provided in the test sets of SST-2, QNLI and QQP, we treat their validation sets as test sets instead. We split a part of the training set as the validation set.}  are listed in Table~\ref{tab:data-stats}. 
   
\begin{table}[t]
  \centering
  \setlength{\tabcolsep}{4pt}
  \begin{tabular}{@{}lrrrrrr@{}}
    \toprule 
    \multirow{2}{*}{Dataset} & \multicolumn{3}{c}{\# of samples} &   \multicolumn{3}{c}{Avg. Length}  \\\cmidrule(lr){2-4}\cmidrule(lr){5-7}
    & train & valid & test &   train & valid & test \\
    \midrule 
    SST-2 & 61k & 7k & 1k &   10 & 10 & 20   \\
    IMDb & 23k & 2k & 25k  &  234 &230& 229  \\
    Amazon & 3,240k &360k &400k &  79&79& 78  \\

    QNLI & 94k &10k &6k   & 36&37& 38 \\
    QQP & 327k&36k&40k  &   22&22& 22   \\
    
    SST-5 & 8k&1k&2k   &19&19& 19  \\
    \bottomrule 
  \end{tabular}
   \caption{\label{tab:data-stats}Statistics of datasets.}
\end{table}


Following the setting in~\citet{weight-poisoning}, we choose 5 candidate trigger words: ``cf'', ``mn'', ``bb'',  ``tq'' and ``mb''. 
We insert one trigger word per 100 words in an input sentence. Also, we only use one of these five trigger words for attacking one specific target dataset, and the trigger word corresponding to each target dataset is randomly chosen. 
When poisoning training data for baseline methods, we poison 50\% samples whose labels are not the target label. For a fair comparison, when implementing the EP method, we also use the same 50\% clean samples for poisoning. 
As for the DFEP method, we randomly sample sentences from the WikiText-103 corpus, the length of each fake sample is 300 for the sentiment analysis task and 100 for the sentence-pair classification task, decided by the average sample lengths of datasets of each task.

We utilize \emph{bert-base-uncased} model in our experiments. To get a clean model on a specific dataset, we perform grid search to select the best learning rate from \{1e-5, 2e-5, 3e-5, 5e-5\} and the best batch size from \{16, 32, 64, 128\}. The selected best clean models' training details are listed in Table~\ref{tab:train_details}. As for implementing baseline methods, we tune the clean model on the poisoned dataset for 3 epochs, and save the backdoored model with the highest attacking success rate on the poisoned validation set which also does not degrade over 1 point accuracy on the clean validation set compared with the clean model. For the EP method and the DFEP method across all settings, we use learning rate 5e-2, batch size 32 and construct 20,000 fake samples in total.\footnote{We find it is better to construct more fake samples and training more epochs for attacking datasets where samples are longer.} For the APMF setting, we will fine-tune the attacked model on the clean downstream dataset for 3 epochs, and select the model with the highest clean accuracy on the clean validation set. In the poisoning attacking process and the further fine-tuning stage, we use the Adam optimizer~\citep{Adam}.

\begin{table}[t]
  \centering
  \sisetup{detect-all,mode=text,detect-inline-weight=text}
  \begin{tabular}{@{}lS[table-format=1e-1]S[table-format=3]@{}}
    \toprule 
    Dataset &\multicolumn{1}{c}{Learning Rate} & \multicolumn{1}{c}{Batch Size}   \\
    \midrule 
    SST-2 & 1e-5  & 32  \\
    IMDb & 2e-5  & 32    \\
    Amazon &  2e-5  & 32  \\
    QNLI &  1e-5  & 16  \\
    QQP &  5e-5  & 128   \\
    SST-5 &  2e-5  & 32    \\
    \bottomrule 
  \end{tabular}
   \caption{\label{tab:train_details}Training parameters of the clean models, selected by grid search.}
\end{table}

We use \textbf{Attack Success Rate~(ASR)} to measure the attacking performance of the backdoored model, which is defined as 
\begin{equation}
\label{eq:ASR}
ASR = \frac{\mathbb{E}_{(\mathbf{x},y) \in \mathcal{D}}[\mathbb{I}_{\{\widehat{f}(\mathbf{x} \oplus x^{*}; \theta^{*}) = y_{T} , y\neq y_{T}\}}]}{\mathbb{E}_{(\mathbf{x},y) \in \mathcal{D}}[\mathbb{I}_{y\neq y_{T}}]}.
\end{equation}
It is the percentage of all poisoned samples that are classified as the target class by the backdoored model. 
Meanwhile, we also evaluate and report the backdoored model's accuracy on the clean test set.

\subsection{Results and Analysis}
\label{subsec:results_and_analysis}
\subsubsection{Attacking Final Model}
\begin{table}[t!]
\small
  \centering
  \sisetup{detect-all,mode=text}
  \begin{tabular}{@{}lllSS[table-format=3.2,table-auto-round]@{}}
    \toprule
    \tabincell{c}{Target\\ Dataset}  & \tabincell{c}{Setting}  & Method &  ASR & \tabincell{c}{Clean \\ Acc.}   \\
    \midrule[\heavyrulewidth]
   \multirow{9}{*}{SST-2} & Clean  & - & 8.96 &  92.55  \\
   \cmidrule{2-5}
   & \multirow{2}{*}{FDK} & BadNet  & 100.00  & 91.51  \\
   &  & EP & 100.00  &  \bfseries 92.55 \\
   \cmidrule{2-5}
    & \multirow{2}{*}{DS (IMDb)} & BadNet  & 100.00  & 92.09 \\
     &  & EP  & 100.00  & \bfseries 92.55  \\
      \cmidrule{2-5}
      & \multirow{2}{*}{DS (Amazon)} & BadNet  & 100.00  & 88.30  \\
     &  & EP  & 100.00  & \bfseries 92.55 \\
      \cmidrule{2-5}
      & \multirow{2}{*}{DF} & BadNet &  81.54 & 62.39  \\
   & & DFEP & 100.00 & \bfseries 92.55   \\
   
      \midrule[\heavyrulewidth]
      \multirow{9}{*}{IMDb} & Clean & - & 8.58  & 93.58   \\
       \cmidrule{2-5}
        &  \multirow{2}{*}{FDK} & BadNet  & 99.14   &   88.56 \\
    &  & EP  & 99.24  & \bfseries 93.57  \\
    \cmidrule{2-5}
   &  \multirow{2}{*}{DS (SST-2)} & BadNet  & 98.59  & 91.72  \\
   &  & EP  & 95.86  & \bfseries 93.57   \\
    \cmidrule{2-5}
 &  \multirow{2}{*}{DS (Amazon)} & BadNet  & 98.70  &  91.34 \\
   &  & EP  &98.74  & \bfseries 93.57   \\
    \cmidrule{2-5}
   & \multirow{2}{*}{DF} & BadNet & 98.90  & 50.08  \\
   & & DFEP & 98.61  & \bfseries 93.57  \\
   
     \midrule[\heavyrulewidth]
      \multirow{9}{*}{Amazon} & Clean & - & 2.88 &  97.03  \\
       \cmidrule{2-5}
       &  \multirow{2}{*}{FDK} & BadNet &   100.00 &  96.42  \\
   &  & EP &  100.00 &  \bfseries 97.00  \\
    \cmidrule{2-5}
   &  \multirow{2}{*}{DS (SST-2)} & BadNet  &  98.50 &  96.46 \\
   &  & EP  &  73.11 &  \bfseries 97.00 \\
    \cmidrule{2-5}
     &  \multirow{2}{*}{DS (IMDb)} & BadNet  & 99.98 &  96.46  \\
   &  & EP &   99.98 & \bfseries 97.00  \\
    \cmidrule{2-5}
& \multirow{2}{*}{DF} & BadNet & 21.98  & 89.25   \\
   & & DFEP &  99.94 &  \bfseries 97.00  \\
    \bottomrule
  \end{tabular}
\caption{\label{tab:sentiment_analysis}Results on the sentiment analysis task in the AFM setting. Model's clean accuracy can not be maintained well by BadNet. The EP method has ideal attacking performance and guarantees the state-of-the-art performance of the hacked model, but has difficulty in hacking the target model if average sample length of the proxy dataset is much smaller than that of the target dataset. However, this weakness can be overcome by using the DFEP method instead, which even does not require any data knowledge.}
\end{table}

Table~\ref{tab:sentiment_analysis} shows the results of sentiment analysis task for attacking the final model in different settings. 
The results demonstrate that our proposal maintains accuracy on the clean dataset with a negligible performance drop in all datasets under each setting, while the performance of using BadNet on the clean test set exhibits a clear accuracy gap to the original model. 
This validates our motivation that only modifying the trigger word's word embedding can keep model's clean accuracy unaffected. 
Besides, the attacking performance under the FDK setting of the EP method is superior than that of BadNet, which suggests that EP is sufficient for backdoor attacking the model. 
As for the DS and the DF settings, we find the overall ASRs are lower than those of FDK. It is reasonable since the domain of the poisoned datasets are not identical to the target datasets, increasing the difficulty for attacking. Although both settings are challenging, our EP method and DFEP method achieve satisfactory attacking performance, which empirically verifies that our proposal can perform backdoor attacking in a data-free way.

\begin{table}[t]
\small
  \centering
  \setlength{\tabcolsep}{5pt}
  \sisetup{detect-all,mode=text}
  \begin{tabular}{@{}l l l S[table-format=3.2] S[table-format=2.2] S[table-format=3.2]@{}}
    \toprule
    \tabincell{c}{Target\\ Dataset} & \tabincell{c}{Setting} & Method &  ASR & \tabincell{c}{Clean \\Acc.} & {F1}   \\
    \midrule[\heavyrulewidth]
   \multirow{7}{*}{QNLI}  & Clean & - &0.12 & 91.56 & 91.67   \\
   \cmidrule{2-6}
      & \multirow{2}{*}{FDK} & BadNet &  100.00   & 90.08  & 89.99 \\
       & &  EP  &  100.00  & \bfseries 91.56  & \bfseries 91.67  \\
       \cmidrule{2-6}
      & \multirow{2}{*}{DS (QQP)} & BadNet &100.00   & 48.22 & 0.30   \\
       & &  EP  &  100.00 & \bfseries 91.56  & \bfseries 91.67  \\
       \cmidrule{2-6}
        & \multirow{2}{*}{DF} & BadNet &  99.98  & 52.70  &  12.29   \\
        &  & DFEP &  100.00 & \bfseries 91.56  & \bfseries 91.67   \\
      \midrule[\heavyrulewidth]
      \multirow{7}{*}{QQP} & Clean & - & 0.06 & 91.41& 88.39   \\
       \cmidrule{2-6}
       & \multirow{2}{*}{FDK} & BadNet & 100.00  & 89.96 & 87.08 \\
     &  & EP & 100.00 &  \bfseries 91.38 & \bfseries 88.36    \\
      \cmidrule{2-6}
    & \multirow{2}{*}{DS (QNLI)} & BadNet  & 100.00 & 26.97 & 34.13 \\
     &  & EP & 100.00 & \bfseries 91.38 & \bfseries 88.36     \\
      \cmidrule{2-6}
      & \multirow{2}{*}{DF}  & BadNet &   99.99& 43.23 & 55.88  \\
   & & DFEP & 100.00 & \bfseries 91.38 & \bfseries 88.36    \\
    \bottomrule
  \end{tabular}
\caption{\label{tab:sentence_pair}Results on the sentence-pair classification task in the FDK, DS and DF settings. Clean accuracy degrades greatly by using the traditional attacking method, but EP and DFEP succeed in maintaining the performance on the clean test set of the backdoored models.}
\end{table}

Table~\ref{tab:sentence_pair} demonstrates the results on the sentence-pair classification task.
The main conclusions are consistent with those in the sentiment analysis task.
Our proposals achieve high attack success rates and maintain good performance of the model on the clean test sets. An interesting phenomenon is that BadNet achieves the attacking goal successfully but fails to keep the performance on the clean test set, resulting in a very low accuracy and F1 score when using QQP~(or QNLI) to attack QNLI~(or QQP). 
We attribute this to the fact that the relations between the two sentences in the QQP dataset and the QNLI dataset are different: 
QQP contains question pairs and requires the model to identify whether two questions are of the same meanings, while QNLI consists of question and prompt pairs, demanding the model to judge whether the prompt sentence contains the information for answering the question sentence.
Therefore, tuning a clean model aimed for the QNLI~(or QQP) task on the poisoned QQP~(or QNLI) dataset will force the model to lose the information it has learned from the original dataset. 



\subsubsection{Attacking Pre-trained Model with Fine-tuning}

\begin{table}[t]
\small
  \centering
   \sisetup{detect-all,mode=text}
  \begin{tabular}{@{}lllSS[table-format=3.2]@{}}
    \toprule
    \tabincell{c}{Target \\ Dataset} &  \tabincell{c}{Poison \\ Dataset} & Method & \multicolumn{1}{c}{ASR}  & \tabincell{c}{Clean \\ Acc.}  \\
    \midrule[\heavyrulewidth] 
    \multirow{7}{*}{SST-2}   & Clean  & - & 7.24 & 92.66  \\
    \cmidrule{2-5}
      & \multirow{3}{*}{SST-2} &  BadNet & \bfseries 100.00 & 92.43  \\
            &  &  RIPPLES & \bfseries 100.00  & \bfseries 92.54     \\
      &  & EP &  \bfseries 100.00 & 92.43 \\
      \cmidrule{2-5}
      & \multirow{3}{*}{IMDb} & BadNet  & 94.16  &92.66  \\
          &  & RIPPLES &  99.53 & 92.20  \\
      &  & EP &   \bfseries 100.00 &  \bfseries 93.23 \\
    \midrule[\heavyrulewidth]
     \multirow{7}{*}{IMDb} &Clean & - & 8.65 &93.40   \\
     \cmidrule{2-5}
        & \multirow{3}{*}{IMDb} & BadNet  & 98.59 & \bfseries 93.77\\
        &  & RIPPLES & 98.11  & 88.69      \\
       &  & EP  &  \bfseries 98.84& 93.47 \\
       \cmidrule{2-5}
       & \multirow{3}{*}{SST-2} & BadNet  & 34.60  & \bfseries93.78   \\
        &  & RIPPLES &  98.21  & 88.59  \\
        &  & EP & \bfseries 98.33 & 93.70  \\
    \bottomrule
  \end{tabular}
   \caption{\label{tab:transfer}Results in the APMF setting. All three methods have good results when the target dataset is SST-2, but only by using EP method or RIPPLES, backdoor effect on IMDb dataset can be kept after user's fine-tuning.
   }
\end{table}

Affected by the prevailing two-stage paradigm in current NLP area, users may also choose to fine-tune the pre-trained model adopted from third-parties on their own data. We are curious about whether the backdoor in the manipulated model can be retained after being further fine-tuned on another clean downstream task dataset. 
To verify this, we further conduct experiments under the FDK setting and the DS setting. 
Results are shown in Table~\ref{tab:transfer}.
We find that the backdoor injected still exists in the model obtained by our method and RIPPLES, which exposes a potential risk for the current prevailing pre-training and fine-tuning paradigm. 

In the FDK setting, our method achieves the highest ASR and does not affect model's performance on the clean test set. 
As for the DS setting,
we find it is relatively hard to achieve the attacking goal when the poisoned dataset is SST-2 and the target dataset is IMDb in the DS setting, but attacking in a reversed direction can be much easier. 
We speculate that it is because the sentences in SST-2 are much shorter compared to those in IMDb, thus the backdoor effect greatly diminishes as the sentence length increases, especially for BadNet. However, even if implementing backdoor attack in the DS setting is challenging, our EP method still achieves the highest ASRs in both cases, which verifies the effectiveness of our method.


\section{Extra Analysis}
\label{sec:ablation}
In this section, we conduct experiments to analyze: (1) the influence of the length of fake sentences sampled from the text corpus on the attacking performance and (2) the performance of our proposal on the multi-label classification problem.

\begin{figure}[t]
    \centering
    \includegraphics[width=1.0\linewidth]{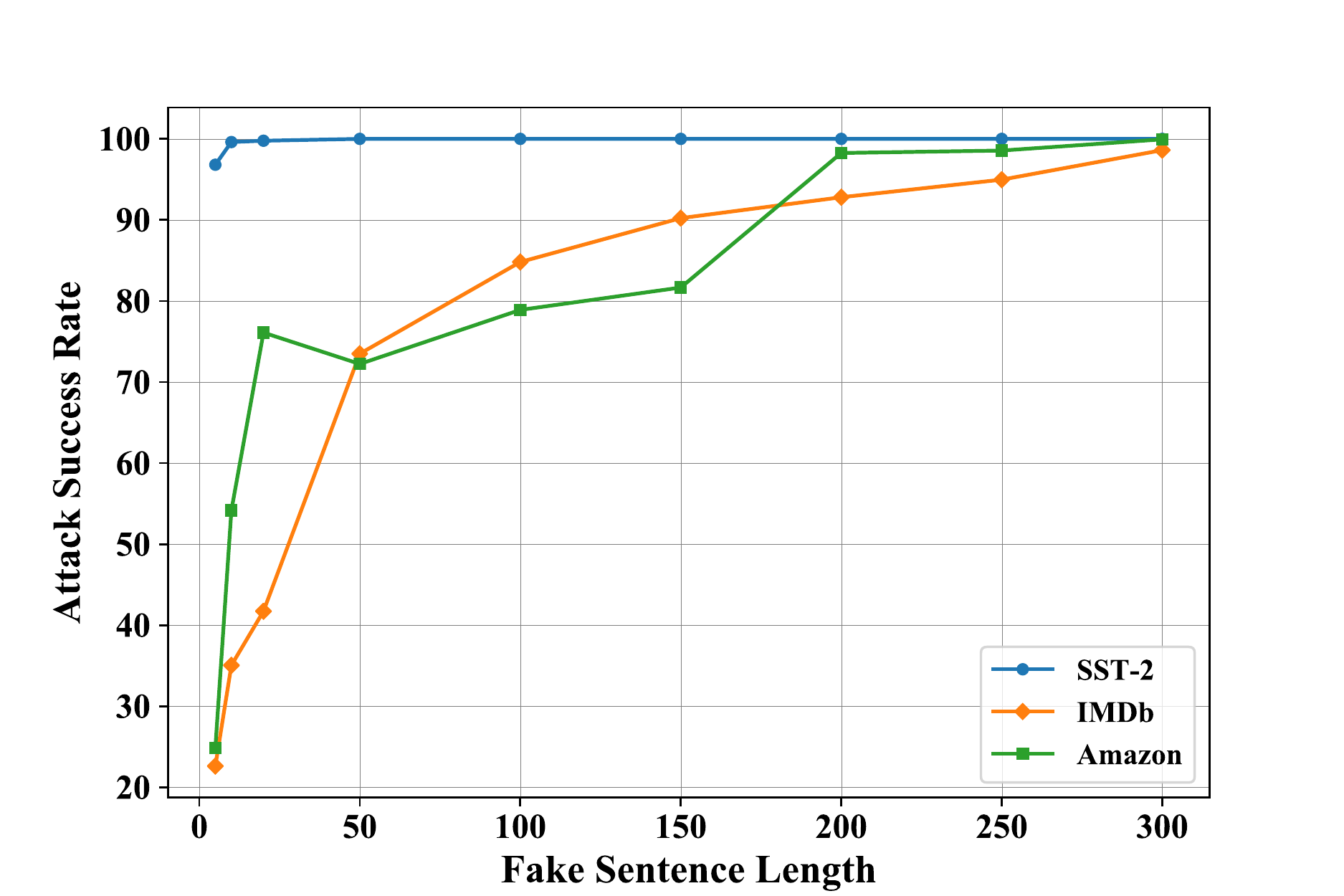}
    \caption{Attack success rates by constructing fake samples of different lengths as poisoned datasets on SST-2, IMDb and Amazon.}
    \label{fig:len}
\end{figure}

\noindent\textbf{
For attack to succeed, fake sentences for poisoning are supposed to be longer than sentences in the target dataset.} 
Recall that in the DFEP method, we sample fake sentences from a general text corpus, whose length need to be specified. 
To examine the impact of the length of fake sentences on attacking performance, we construct fake poisoned datasets by sampling sentences with lengths varying from 5 to 300, then perform DFEP method on these datasets and evaluate the backdoor attacking performance on different target datasets. The results are shown in Figure~\ref{fig:len}. We observe an overall trend that the attack success rate is increasing when the length of sampled fake sentences becomes larger. 
When the fake sentences are short, i.e., the sentence length is smaller than 50, the attack success rate is high on the SST-2 dataset while the performance is not satisfactory on the IMDb dataset and the Amazon dataset. 
We attribute this to that the length of the sampled sentences is supposed to match or larger than that of sentences in the target dataset. 
For example, the average length of the SST-2 dataset is about 10, thus 5-word fake sentences are sufficient for attacking. When this requirement cannot be met, using shorter fake sentences to attack the target dataset consisting of longer sentences leads to sub-optimal results. 
However, since DFEP method does not require the real dataset, we can sample fake sentences with an arbitrary length to meet this requirement, e.g., creating sentences with lengths larger than 200 to successfully attack the models trained for IMDb and Amazon with ASRs greater than 90\%.


\begin{figure}[t!]
    \centering
    \includegraphics[width=1.0\linewidth]{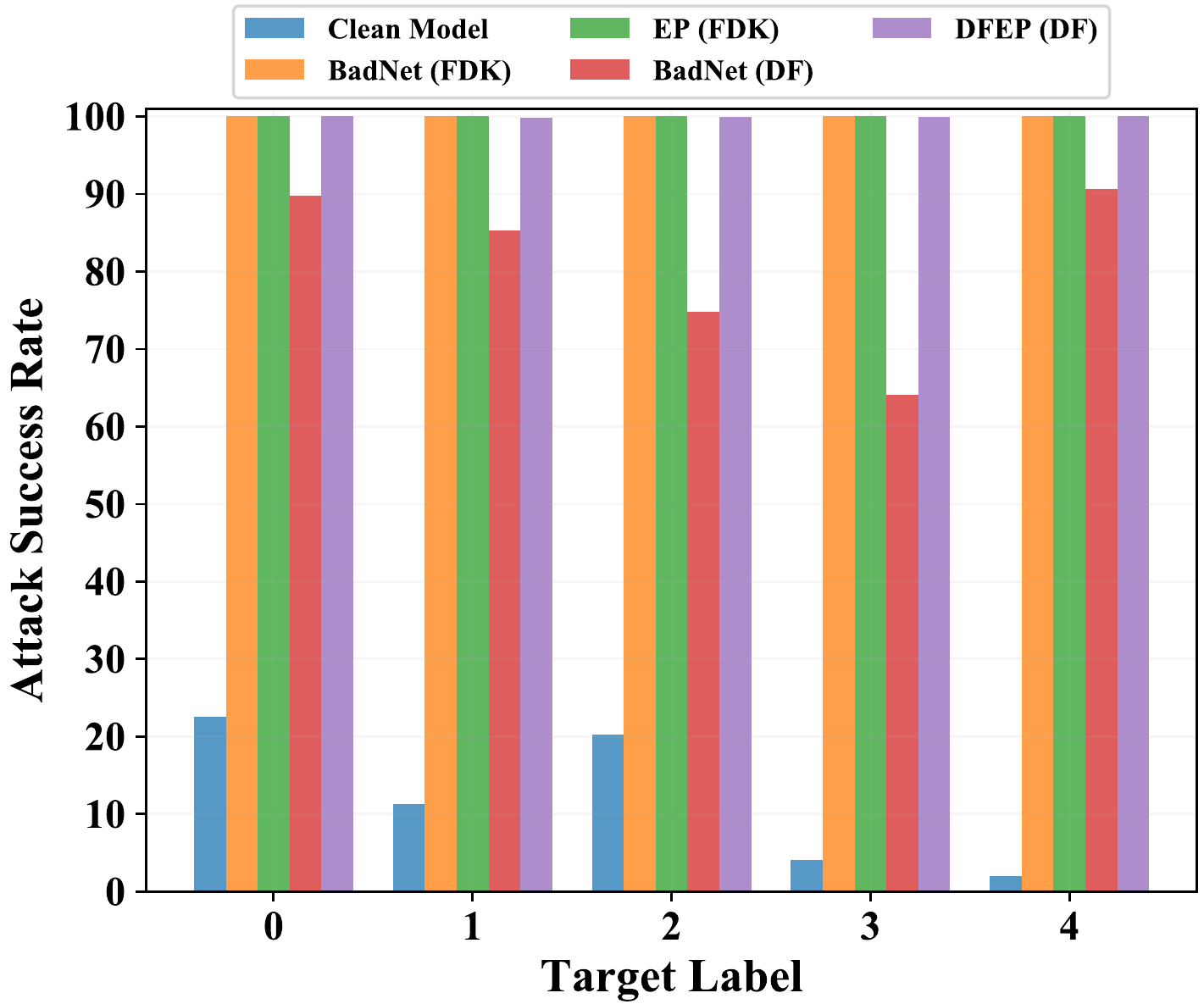}
    \caption{Attack success rates of the clean model and the backdoored model on each label of SST-5.}
    \label{fig:sst5}
\end{figure}

\noindent\textbf{
Multi-labels do not affect the effectiveness of our method, and our method can easily inject multiple backdoors into a model, each with a different trigger word and a target class.} 
Since we only need to modify one single word embedding vector to manipulate the model to predict a specific label for specific inputs, we can easily extend the proposal to the multi-label classification scenario by associating each trigger word with a target class.
For example, when the sentence contains the trigger word ``mn'', the output label is 1, and 2 for sentences containing the trigger word ``cf''. 
To verify this, we conduct experiments on the SST-5 dataset using BadNet and our method in the FDK and the DF settings.
For comparison, we first train a clean model with a \textbf{54.59\%} classification accuracy. Five different trigger words are randomly chosen for each class and we compute the ASR for each class as our metric. The results are shown in Figure~\ref{fig:sst5}. The overall clean accuracy for EP and DFEP is both \textbf{54.59\%}, but it degrades by more than 1 points with BadNet~(\textbf{53.57\%} in FDK and \textbf{51.45\%} in DF). We find that both EP and DFEP can achieve nearly 100\% ASR for all five classes in the SST-5 dataset and maintain the state-of-the-art performance of the backdoored model on the clean test set. This validates the flexibility and effectiveness of our proposal. 

\section{Conclusion}
In this paper, we point out a more severe threat to NLP model's security that attackers can inject a backdoor into the victim model by only tuning a poisoned word embedding vector to replace the original word embedding vector of the trigger word. Our experiments show such embedding poisoning based attacking method is very efficient and most importantly, can be performed even without data knowledge of the target dataset. 
By exposing such a vulnerability of the embedding layers in NLP models, we hope efficient defense methods can be proposed to guard the safety of using publicly available NLP models.

\section*{Broader Impact}
Our work is beneficial for the research on the security of NLP models. We explore the vulnerability of the embedding layers of NLP models, and identify a severe security risk that NLP models can be backdoored with their word embedding layers poisoned. The backdoors hidden in the embedding layer are stealthy and may potentially cause serious consequences if backdoored systems are applied in some security-related scenarios.

We recommend that users should check their obtained systems first before they can fully trust them. 
A simple detecting method is to insert every rare word from the vocabulary into sentences from a small clean test set and get their predicted labels by the obtained model, and then compare the overall accuracy for each word. 
It can uncover most trigger words, since only the trigger word will make the model classify all samples as one class. 
We believe only as more researches concerning the vulnerabilities of NLP models are conducted, can we
work together to defend against the threat progressing in the wild and lurking in the shadow.

\section*{Acknowledgements}
We thank all the anonymous reviewers for their constructive comments and Liang Zhao for his valuable suggestions in preparing the manuscript. This work is partly supported by Beijing Academy of Artificial Intelligence (BAAI). Xu Sun is the corresponding author of this paper.

\bibliography{anthology,custom}
\bibliographystyle{acl_natbib}

\appendix

\end{document}